\documentclass[11pt,a4paper]{article}
\usepackage[hyperref]{acl2019}
\usepackage{times}
\usepackage{latexsym}
\usepackage{graphicx,subfigure}
\usepackage{url}

\aclfinalcopy % Uncomment this line for the final submission
\def\aclpaperid{2432} %  Enter the acl Paper ID here

\title{A Corpus for Modeling User and Language Effects\\ in Argumentation on Online Debating}

\author{Esin Durmus \\
  Cornell University \\
  \texttt{ed459@cornell.edu} \\\And
  Claire Cardie \\
  Cornell University \\
  \texttt{cardie@cs.cornell.edu}
  }

\date{}

\begin{document}
\maketitle
\begin{abstract}
Existing argumentation datasets have succeeded in allowing researchers 
to develop computational 
methods for analyzing the content, structure and linguistic features of argumentative
text. 
They have been much less successful in fostering studies of the effect of 
``user" traits --- characteristics and beliefs of the participants --- 
on the debate/argument outcome as this type of user information is generally
not available.  This paper presents a dataset of $78,376$ debates generated
over a 10-year period along
with surprisingly comprehensive participant profiles. 
We also complete an example study using the dataset
to analyze the effect of selected user traits 
on the debate outcome in comparison to the
linguistic features typically employed in studies of this kind. 
\end{abstract}
\section{Introduction}
Previous work from Natural Language Processing (NLP) and Computational
Social Science (CSS) that studies argumentative text and its persuasive
effects has mainly focused on identifying the content and structure of
an argument (e.g. \newcite{feng2011classifying}) and the linguistic features that
are indicative of effective argumentation strategies (e.g.\ \newcite{DBLP:journals/corr/TanNDL16}).
The effectiveness of an argument, however, cannot be determined solely
by its textual content; rather, it is important to
consider characteristics of the reader, listener or participants
in the debate or discussion.  Does the reader already agree with the argument's
stance?  Is she predisposed to changing her mind on the particular topic of
the debate?  Is the style of the argument appropriate for
the individual? To date, existing argumentation datasets have permitted only
limited assessment of such ``user" traits because information on the
background of users is generally unavailable.
In this paper, we present a dataset of  $78,376$ debates from October of
2007 until November of 2017 drawn from \href{http://www.debate.org}{\it debate.org} along with quite
comprehensive user profile information --- for debate participants as well as
users voting on the debate quality and outcome.  Background
information on users includes demographics (e.g.\ education, income,
religion) and stance on a variety of controversial debate topics 
as well as a
record of user activity on the debate platform (e.g.\ debates won and
lost). We view this new dataset as a resource that affords the NLP
and CSS communities the opportunity to understand the effect of
audience characteristics on the efficacy of different debating and
persuasion strategies as well as to model changes in user's opinions
and activities on a debate platform over time. (To date, part of our \href{http://www.debate.org}{\it debate.org} dataset has been used in one such study to understand the effect of prior beliefs in persuasion\footnote{That study is distinct from those presented here. See Section \ref{task_section} for details.} \cite{durmus-cardie-2018-exploring}.  
Here, we focus on the properties of the dataset itself and study a different task.)

In the next section, we describe the dataset in the context of
existing argumentation datasets. We then provide statistics on key
aspects of the collected debates and user profiles
(Section~\ref{statistics_section}).
Section~\ref{task_section} reports a study in which we investigate 
the predictive effect of selected user traits (namely, the debaters' and audience's 
experience, prior debate success, social interactions, and demographic information) vs.\ standard linguistic features. Experimental results show that features of the user traits are significantly more predictive of a debater's success than the linguistic features that are shown to be predictive of debater success by the previous work \cite{zhang2016conversational}. This suggests that user traits are important to take into account in studying success in online debating.  

The dataset will be made publicly
available\footnote{Link to the dataset: \href{http://www.cs.cornell.edu/~esindurmus/}{http://www.cs.cornell.edu/~esindurmus/}.}.
\begin{figure*}
\centering
\includegraphics[scale=0.46]{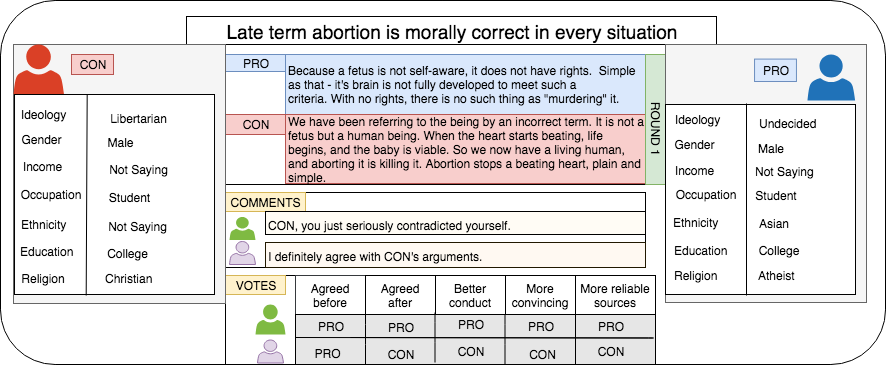}
\caption{Example debate along with the user profile information for {\sc pro} and {\sc con} debaters and the corresponding comments and votes. The full information for this debate can be found at \href{https://www.debate.org/debates/Late-term-abortion-is-morally-correct-in-every-situation/1/comments/1/}{https://www.debate.org/debates/Late-term-abortion-is-morally-correct-in-every-situation/1/}.}
\label{example_debate}
\end{figure*}

\section{Related Work and Datasets}
There has been a tremendous amount of research effort to understand the important linguistic 
features for identifying argument structure and determining effective argumentation strategies
in monologic
text \citep{article_moens,feng2011classifying,Stab2014IdentifyingAD,article}. For example, \newcite{habernal2016makes} has experimented with different machine learning models to predict 
which of two arguments is more convincing. To understand what kind of persuasive strategies are 
effective, \newcite{W17-5102} has further annotated different modes of persuasion (ethos, logos, 
pathos) and looked at which combinations appear most often in more persuasive arguments.  
% \begin{table*}
% \begin{center}
% \begin{tabular}{p{3.5cm}p{12.5cm}}
% \hline
% \textbf{Category.}&\textbf{Information} \\ 
% \hline
% \textbf{Demographic.} &  Ideology, Gender, Birthday, Education, Ethnicity, Income, Occupation, Religion.\\\hline
% \textbf{Stance.} &  Stance information ({\sc pro/con}, {\sc no opinion}, {\sc not saying}, {\sc undecided}) for 48 controversial debate topics such as \textit{{\sc abortion} and {\sc death penalty}}. \\\hline
% \textbf{Activity.} & Time that the user joined to the community, time of last activities. \\\hline
% \textbf{Other views.} & Opinion questions, opinion arguments, poll votes, poll questions of the users. \\\hline
% \textbf{Debate success.} &  Fraction of debates the user won, overall ranking of the user on the website.\\\hline \\ 
% \textbf{Social Network.} & Friends of the user, the opponents of the user in the participated debates.
% \\\hline
% \end{tabular}
% \end{center}
% \caption{Available user information}
% \label{table:1}
% \end{table*}

Understanding argumentation strategies in conversations and the effect of interplay between the language of the participants has also been an important avenue of research. \newcite{DBLP:journals/corr/TanNDL16}, for example,  has examined the effectiveness of arguments
on ChangeMyView\footnote{\href{https://www.reddit.com/r/changemyview/} {https://www.reddit.com/r/changemyview/}.}, a debate forum website  in which people invite others
to challenge their opinions. They found that the interplay between the language of the opinion holder and that of the counterargument provides highly predictive cues of persuasiveness. \newcite{zhang2016conversational} has examined the effect of conversational style in Oxford-style debates and found that the side that can best adapt in response to opponents' discussion points over the course of the debate is more likely to be more persuasive.   
Although research on computational argumentation has mainly focused on identifying important
linguistic features 
%for argumentation strategies looking at only 
of the text, there is also evidence that it is important to model 
%and account for the information about 
the debaters themselves and the people who are judging the quality of the arguments: 
multiple studies show that people perceive arguments from different perspectives depending on their backgrounds and experiences \cite{correll2004affirmed,hullett2005impact,petty1981personal,lord1979biased,vallone1985hostile,doi:10.1177/0741088396013003001}. 
As a result, we introduce data from a social media debate site %devoted to argumentation.  All arguments are in the form of a multiple round debate.   and the 
that also includes substantial information about its users and their activity and interaction on the website.  This is in contrast to the datasets commonly employed in studies of argument strategies \cite{4243c21af1a5470a82c9ec69b9af1e90,WALKER12.1078.L12-1643,zhang2016conversational,Wang2017WinningOT,N16-1166,C16-1324}. \newcite{lukin2017argument} is the closest work to ours as it studies the effect of {\sc ocean} personality traits \cite{doi:10.1177/0146167202289008,article2} of the audience on how they perceive the persuasiveness of monologic arguments. Note that, in our dataset, we do not have information about users' personality traits; however, we have extensive information about their demographics, social interactions, beliefs and language use.

\section{Dataset\footnote{Data is crawled in accordance to the terms and conditions of the website.}}\label{statistics_section}
\textbf{Debates.} The dataset includes $78,376$ debates from $23$ different topic categories including \textit{Politics}, \textit{Religion}, \textit{Technology}, \textit{Movies}, \textit{Music}, \textit{Places-Travel}. Each debate consists of different rounds in which opposing sides provide their arguments. An example debate along with the user information for {\sc pro} and {\sc con} debaters and corresponding comments and votes are shown in Figure \ref{example_debate}. The majority of debates have three or more rounds; \textit{Politics}, \textit{Religion}, and \textit{Society} are the most common debate categories. Each debate includes comments as well as the votes provided by other users in the community. We collected all the comments and votes for each debate with 606,102 comments and 199,210 votes in total. Voters evaluate each debater along diverse set of criteria such as convincingness, conduct during the debate, reliability of resources cited, spelling and grammar. With this fine-grained evaluation scheme, we  can study the quality of arguments from different perspectives.

\textbf{User Information.} The dataset also includes self-identified information for $45,348$ users participating in the debates or voting for the debates: demographic information such as age, gender, education, ethnicity; prior belief and personal information such as political, religious ideology, income, occupation and the user's stance on a set of $48$ controversial topics chosen by the website. The controversial debate topics\footnote{Full list of topics: \href{https://www.debate.org/big-issues/}{https://www.debate.org/big-issues/}.} include {\sc abortion}, {\sc death penalty}, {\sc gay marriage}, and {\sc affirmative action}.
Information about user's activity is also provided and includes their debates, votes, comments, opinion questions they ask, poll votes they participated in, overall success in winning debates as well as their social network information.
\section{Task: What makes a debater successful?}\label{task_section}
To understand the effect of user characteristics vs.\ language features, and staying consistent with majority of previous work, we conduct the task of predicting the winner of a debate by looking at accumulated scores from the voters. We model this as a binary classification task and experiment with a logistic regression model, optimizing the regularizer ($\ell$1 or $\ell$2) and the
regularization parameter C (between $10^{-5}$ and $10^{5}$) with 3-fold cross validation. 
\subsection{Data preprocessing}
\textbf{Controlling for the debate text.} We eliminate debates where a debater forfeits before the debate ends. From the remaining debates, we keep only the ones with three or more rounds with at least 20 sentences by each debater in each round to be able to study the important linguistic features \footnote{After all the eliminations, we have 1635 debates in our dataset.}. \\
\textbf{Determining the winner.} For this particular dataset, the winning debater is determined by the votes of other users on different aspects of the arguments as outlined in Section \ref{statistics_section}, and the debaters are scored accordingly\footnote{Having better conduct: 1 point, having better spelling and grammar: 1 point, making more convincing arguments: 3 points, using the most reliable sources: 2 points.}. We determine the winner by the total number of points the debaters get from the voters. We consider the debates with at least $5$ voters and remove the debates resulting in a tie.
%  \textbf{Controlling for user characteristics.} Since we want to study the effect of user traits, we look at the debates where debaters have different political ideologies\footnote{We eliminate the debates where at least one of the debaters prefer not to share their political ideology.}. We chose this setting so that we can eliminate debates where the debaters have similar user profiles, in order to study whether similarity in user traits between voters and debaters can influence the outcome of a debate. 
%  From these debates, we keep the ones having at least five voters who share their political ideologies. \\

\subsection{Features}
 
\textbf{Experience and  Success Prior.} We define the \textbf{experience} of a user during a debate \textit{$d_t$} at time \textit{t} as the total number of debates participated as a debater by the user before time \textit{t}. The \textbf{success prior} is defined as the ratio of the number of debates the user won before time \textit{t} to the total number of debates before time \textit{t}. 

\textbf{Similarity with audience's user profile.}\label{similarity_a_d}
We encode the similarity of each of the debaters and the voters by comparing each debaters' opinions on controversial topics, religious ideology, genders, political ideology, ethnicity and education level to same of the audience. We include the features that encode the similarity by counting number of voters having the same values as each of the debaters for each of these characteristics. We also include features that corresponds to cosine distance between the vectors of each debater and each voter where the user vector is one-hot representation for each user characteristic.

\textbf{Social Network.} We extract features that represent the debaters' social interactions before a particular debate by creating the network for their commenting and voting activity before that debate. We then computed the degree, centrality, hub and authority scores from these graphs and include them as features in our model.

\textbf{Linguistic features of the debate.} 
We perform ablation analysis with various linguistic features shown to be effective in determining persuasive arguments including argument lexicon features \cite{somasundaran2007detecting}, politeness marks \cite{P13-1025}, sentiment, connotation \cite{feng2011classifying}, subjectivity \cite{wilson2005recognizing}, modal verbs, evidence (marks of showing evidence including words and phrases like ``evidence'' ,``show'',
``according to'', links, and numbers), hedge words \cite{tan+lee:16}, positive words, negative words, swear words, personal pronouns, type-token ratio, tf-idf, and punctuation. To get a text representation for the debate, we concatenated all the turns of each of the participants, extracted features for each and finally concatenated the feature representation of each participant's text.\\
We also experimented with \textit{conversational flow features} shown to be effective in determining the successful debaters by \cite{zhang2016conversational} to track how ideas flow between debaters throughout a debate. Consistent with \cite{zhang2016conversational}, to extract these features, we determine the \textit{talking points} that are most discriminating words for each side from the first round of the debate applying the method introduced by \cite{monroe_colaresi_quinn} which estimates the divergence between
the two sides’ word-usage.\\

\begin{table}[tbp]
\centering
\begin{tabular}{ |l|c|}
 \hline
  & Accuracy \\
 \hline
 Majority baseline & 57.23 \\
 \hline
 \textbf{User features} & \\ 
 \hline
%  \hline
 Debate experience & 63.54 \\ 
 \hline
 Success prior & 65.78 \\ 
 \hline
 Overall similarity with audience & 62.52 \\ 
 \hline
 Social network features & 62.93 \\
\hline
 All user features & \textbf{68.43} \\ 
 \hline
%  \hline
 \textbf{Linguistic features} & \\ 
 \hline
Length& $58.45$ \\
\hline 
Flow features & 58.66 \\
\hline
All linguistic features & \textbf{60.28} \\ 
\hline
\textbf{User+Linguistic Features} & \textbf{71.35}\\ 

% Flow features & $57.94$ \\
% All linguistic features & $58.53$\\

\hline
\end{tabular}
\caption{Ablation tests for the features. }
\label{language_result_table}
\end{table} 

\subsection{Results and Analysis}
Table \ref{language_result_table} shows the results for the user and linguistic features. We find that combination of the debater experience, debater success prior, audience similarity features and debaters' social network features performs significantly better\footnote{We measure the significance performing t-test.}  than the majority baseline and linguistic features achieving the best accuracy ($68.43$\%). We observe that experience and social interactions are positively correlated with success. It suggests that as debaters spend more time on the platform, they probably learn strategies and adjust to the norms of the platform and this helps them to be more successful. We also find that success prior is positively correlated with success in a particular debate. In general, the debaters who win the majority of the debates when first join the platform, tend to be successful in debating through their lifetime. This may imply that some users may already are good at debating or develop strategies to win the debates when they first join to the platform. Moreover, we find that similarity with audience is positively correlated with success which shows that accounting for the characteristics of the audience is important in persuasion studies \cite{lukin2017argument}.  

Although the linguistic features perform better than the majority baseline, they are not able to achieve as high performance as the features encoding debater and audience characteristics. This suggest that success in online debating may be more related to the users' characteristics and social interactions than the linguistic characteristics of the debates. We find that use of argument lexicon features and subjectivity are the most important features and positively correlated with success whereas \textit{conversational flow features} do not perform significantly better than length. This may be because debates in social media are much more informal compare to Oxford style debates and therefore, in the first round, the debaters may not necessarily present an overview of their arguments (\textit{talking points}) they make through the debate. 

We observe that (44\%) of the mistakes made by the model with user features are classified correctly by the linguistic model. This motivated us to combine the user features with linguistic features which gives the best overall performance (71.35\%). This suggests that user aspects and linguistic characteristics are both important components to consider in persuasion studies. We believe that these aspects complement each other and it is crucial to account for them to understand the actual effect of each of these components. For future work, it may be interesting to understand the role of these components in persuasion further and think about the best ways to combine the information from these two components to better represent a user.
\section*{Acknowledgments} 
This work was supported in part by NSF grants IIS-1815455 and SES-1741441.  The views and conclusions contained herein are those of the authors and should not be interpreted as necessarily representing the official policies or endorsements, either expressed or implied, of NSF or the U.S.\ Government.
%  \section{Conclusion}
% In this paper, we focus on the persuasion aspect of argumentation considering linguistic features, debater profiles and the audience profile at the same time. We show that debater and audience features are more predictive for our study than the linguistic features typically employed in predicting the successful debater. We believe that this dataset will open new avenues in argumentation research as it provides the opportunity to study the effect of audience characteristics, debater characteristics and the language of the debate to determine success in argumentation. 
\bibliography{acl2019}
\bibliographystyle{acl_natbib}

\end{document}